\documentclass[conference]{IEEEtran}
\IEEEoverridecommandlockouts
\usepackage{cite}
\usepackage{amsmath,amssymb,amsfonts}
\usepackage{algorithm}
\usepackage{algorithmicx}
\usepackage{algpseudocode}
\usepackage{graphicx}
\usepackage{textcomp}
\usepackage{xcolor}
\usepackage{booktabs}
\usepackage{multirow}
\usepackage{bm}
\usepackage{subfigure} 
\makeatletter
\newcommand{\removelatexerror}{\let\@latex@error\@gobble}
\makeatother

\def\BibTeX{{\rm B\kern-.05em{\sc i\kern-.025em b}\kern-.08em
    T\kern-.1667em\lower.7ex\hbox{E}\kern-.125emX}}
\begin{document}

\title{A Human Eye-based Text Color Scheme Generation Method for Image Synthesis \\
}

\author{\IEEEauthorblockN{
		Shaowei Wang\IEEEauthorrefmark{1},	
		Guanjie Huang\IEEEauthorrefmark{2}, 
		Xiangyu Luo\IEEEauthorrefmark{3}
	}
\IEEEauthorblockA{College of Computer Science and Technology, Huaqiao University, Xiamen, China\\
		Email: 
		\IEEEauthorrefmark{1}jem2rsc@gmail.com,
		\IEEEauthorrefmark{2}wbshgj@163.com,
		\IEEEauthorrefmark{3}luoxy@hqu.edu.cn \\
		Corresponding Author: Xiangyu Luo,     
		Email: luoxy@hqu.edu.cn }
}

\maketitle

\begin{abstract}
Synthetic data used for scene text detection and recognition tasks have proven effective. However, there are still two problems: First, the color schemes used for text coloring in the existing methods are relatively fixed color key-value pairs learned from real datasets. The dirty data in real datasets may cause the problem that the colors of text and background are too similar to be distinguished from each other. Second, the generated texts are uniformly limited to the same depth of a picture, while there are special cases in the real world that text may appear across depths. To address these problems, in this paper we design a novel method to generate color schemes, which are consistent with the characteristics of human eyes to observe things. The advantages of our method are as follows: (1) overcomes the color confusion problem between text and background caused by dirty data; (2) the texts generated are allowed to appear in most locations of any image, even across depths; (3) avoids analyzing the depth of background, such that the performance of our method exceeds the state-of-the-art methods; (4) the speed of generating images is fast, nearly one picture generated per three milliseconds. The effectiveness of our method is verified on several public datasets.
\end{abstract}

\begin{IEEEkeywords}
	color scheme, human visual characteristics, synthesis images
\end{IEEEkeywords}

\section{Introduction}
With the emergence of synthetic data, it effectively solves the bottleneck problem caused by model data-thirsty. The current study widely uses it to train models for the advantages of not requiring manual labeling, saving time, effort, cost, and so on. Several synthesis algorithms~\cite{b17,b21,b30,b38} have made great contributions to the development of synthetic data. Represented by~\cite{b11} and~\cite{b38} successfully dealt with the problem of text recognition.~\cite{b2,b6,b7,b13,b15} were dedicated to solving text detection.

However, two problems are easily overlooked in the existing methods: First, the color scheme used for text rendering learns from the real dataset, which contains some inevitable dirty data. The result is that the text and the background are difficult to distinguish. For
\begin{figure}[ht]
	\centering
	\subfigure[]{
		\label{Fig.sub.1}
		\includegraphics[width=0.22\columnwidth,height=0.08\textheight]{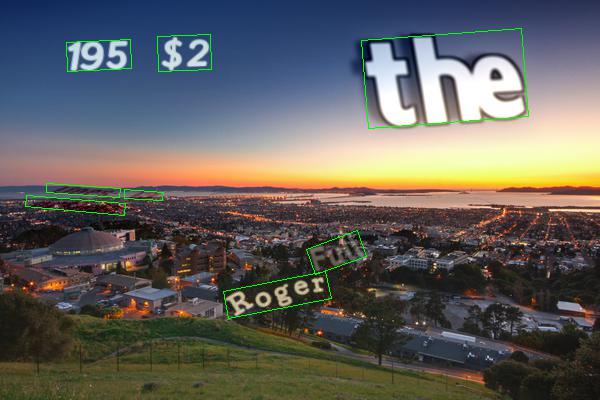}}
	\subfigure[]{
		\label{Fig.sub.2}
		\includegraphics[width=0.22\columnwidth,height=0.08\textheight]{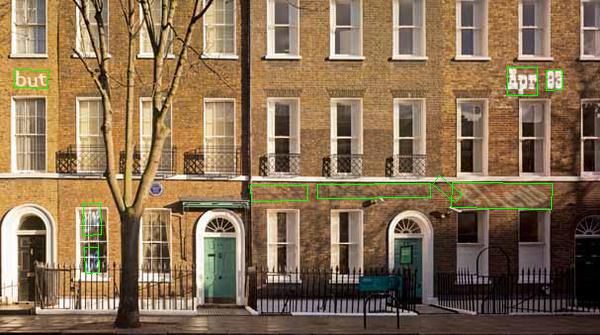}}
	\subfigure[]{
		\label{Fig.sub.3}
		\includegraphics[width=0.22\columnwidth,height=0.08\textheight]{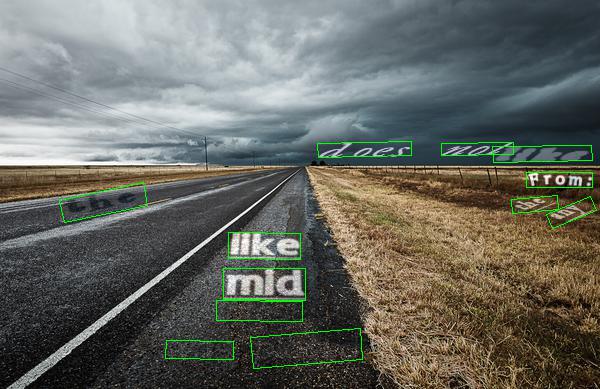}}
	\subfigure[]{
		\label{Fig.sub.4}
		\includegraphics[width=0.22\columnwidth,height=0.08\textheight]{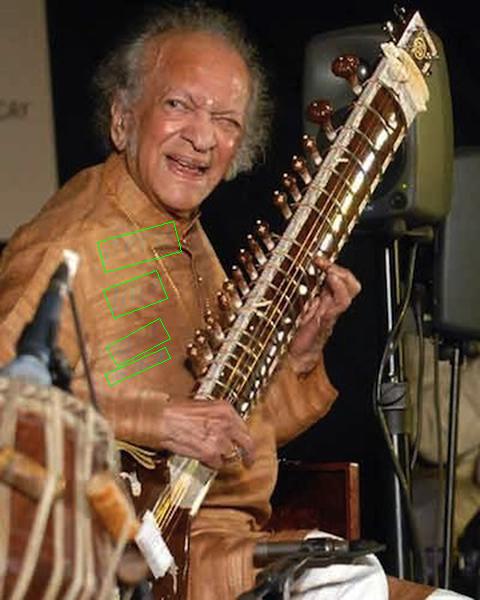}}
	\subfigure[]{
		\label{Fig.sub.5}
		\includegraphics[width=0.22\columnwidth,height=0.08\textheight]{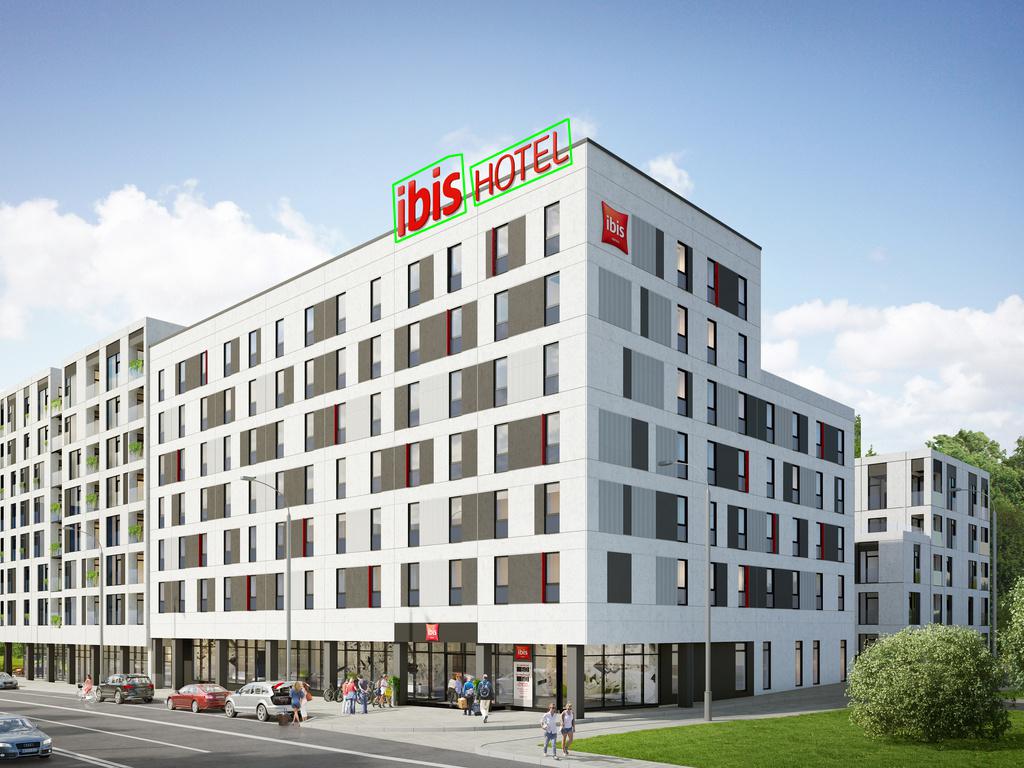}}
	\subfigure[]{
		\label{Fig.sub.6}
		\includegraphics[width=0.22\columnwidth,height=0.08\textheight]{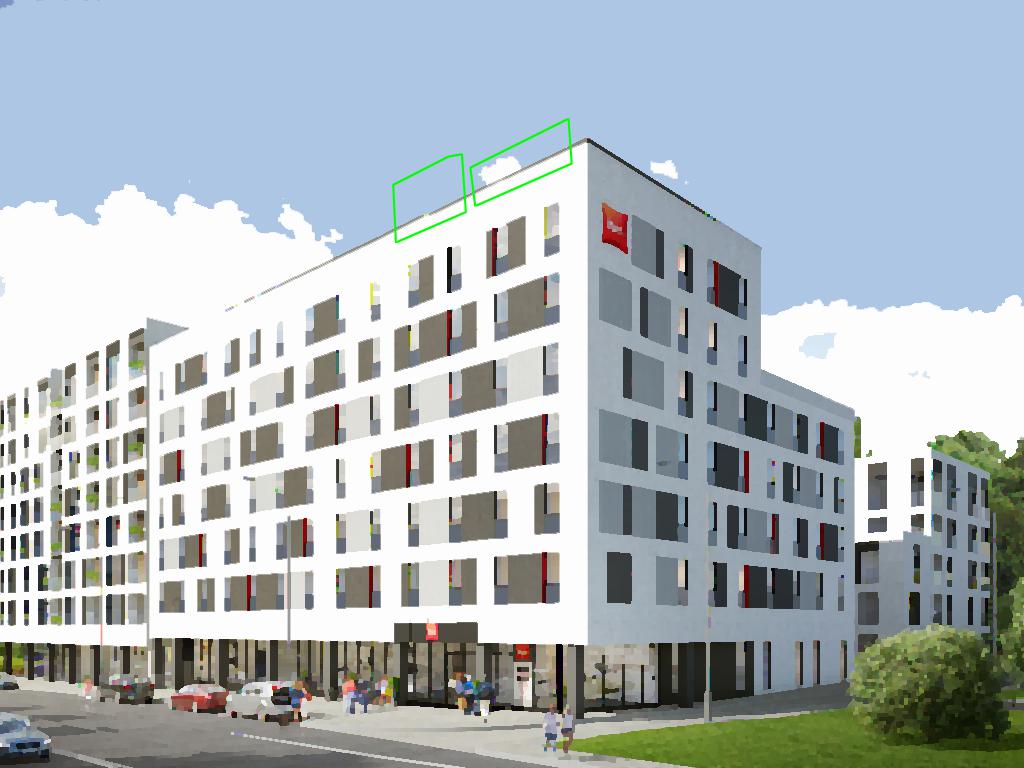}}
	\subfigure[]{
		\label{Fig.sub.7}
		\includegraphics[width=0.22\columnwidth,height=0.08\textheight]{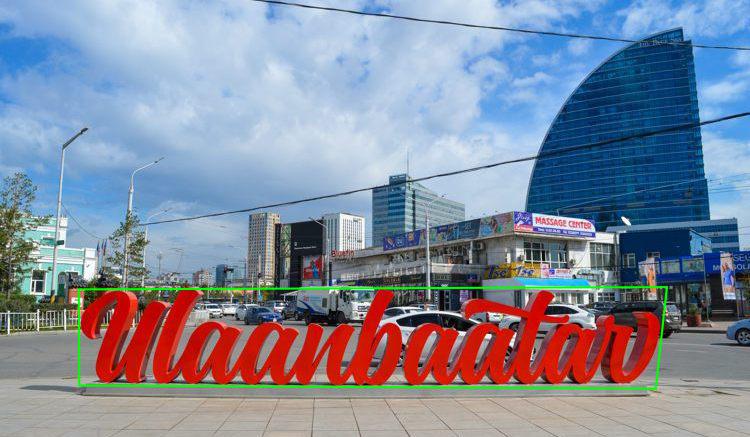}}
	\subfigure[]{
		\label{Fig.sub.8}
		\includegraphics[width=0.22\columnwidth,height=0.08\textheight]{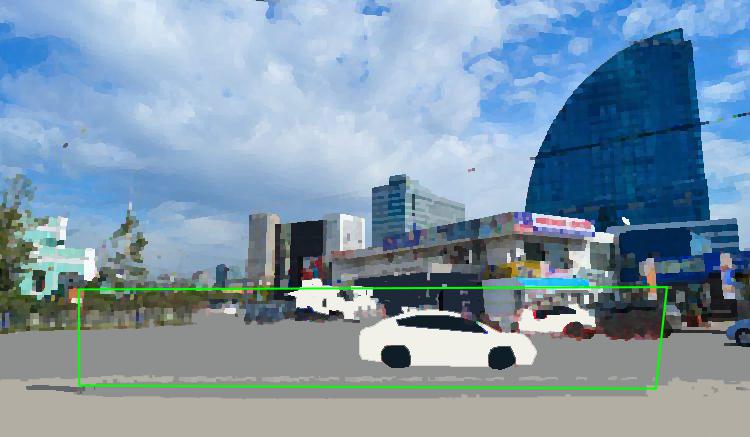}}	
	\subfigure[]{
		\label{Fig.sub.13}
		\includegraphics[width=0.22\columnwidth,height=0.08\textheight]{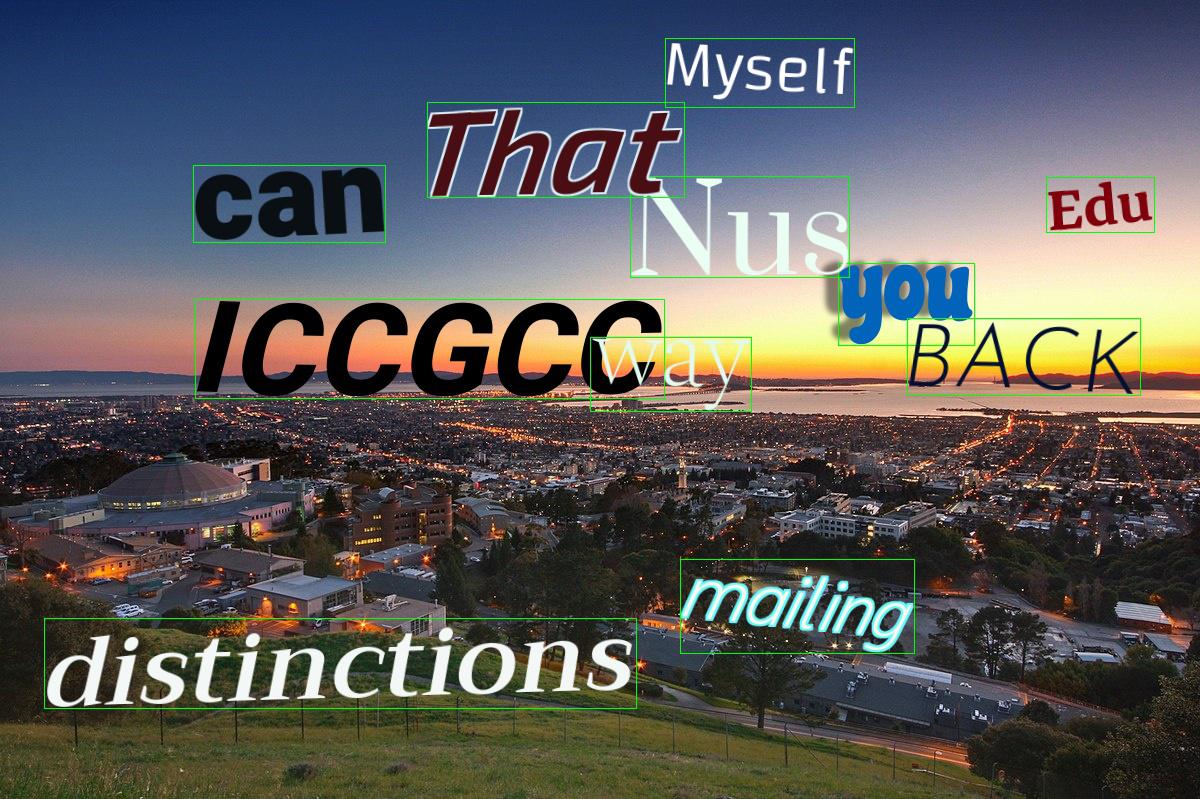}}
	\subfigure[]{
		\label{Fig.sub.14}
		\includegraphics[width=0.22\columnwidth,height=0.08\textheight]{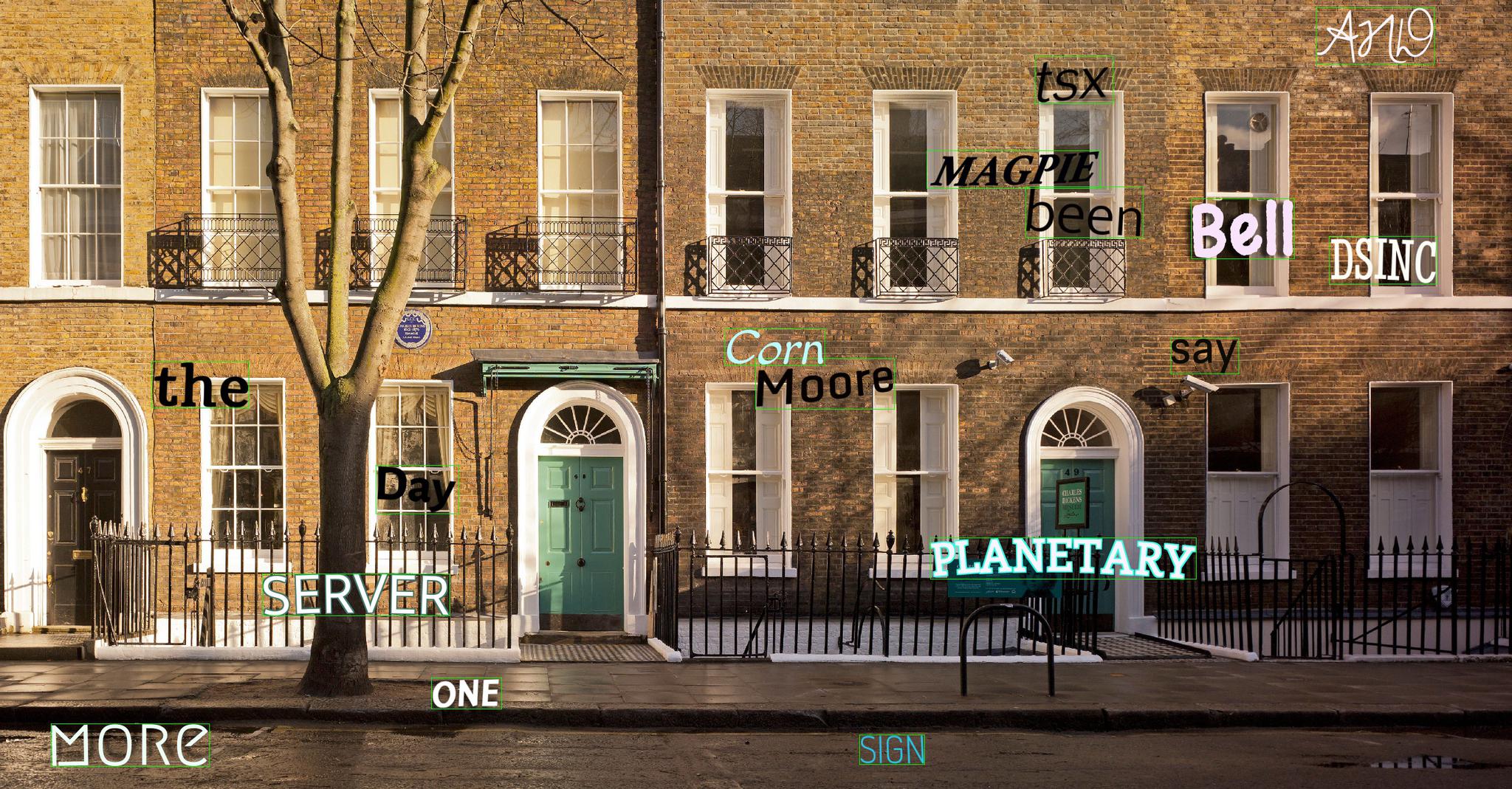}}
	\subfigure[]{
		\label{Fig.sub.15}
		\includegraphics[width=0.22\columnwidth,height=0.08\textheight]{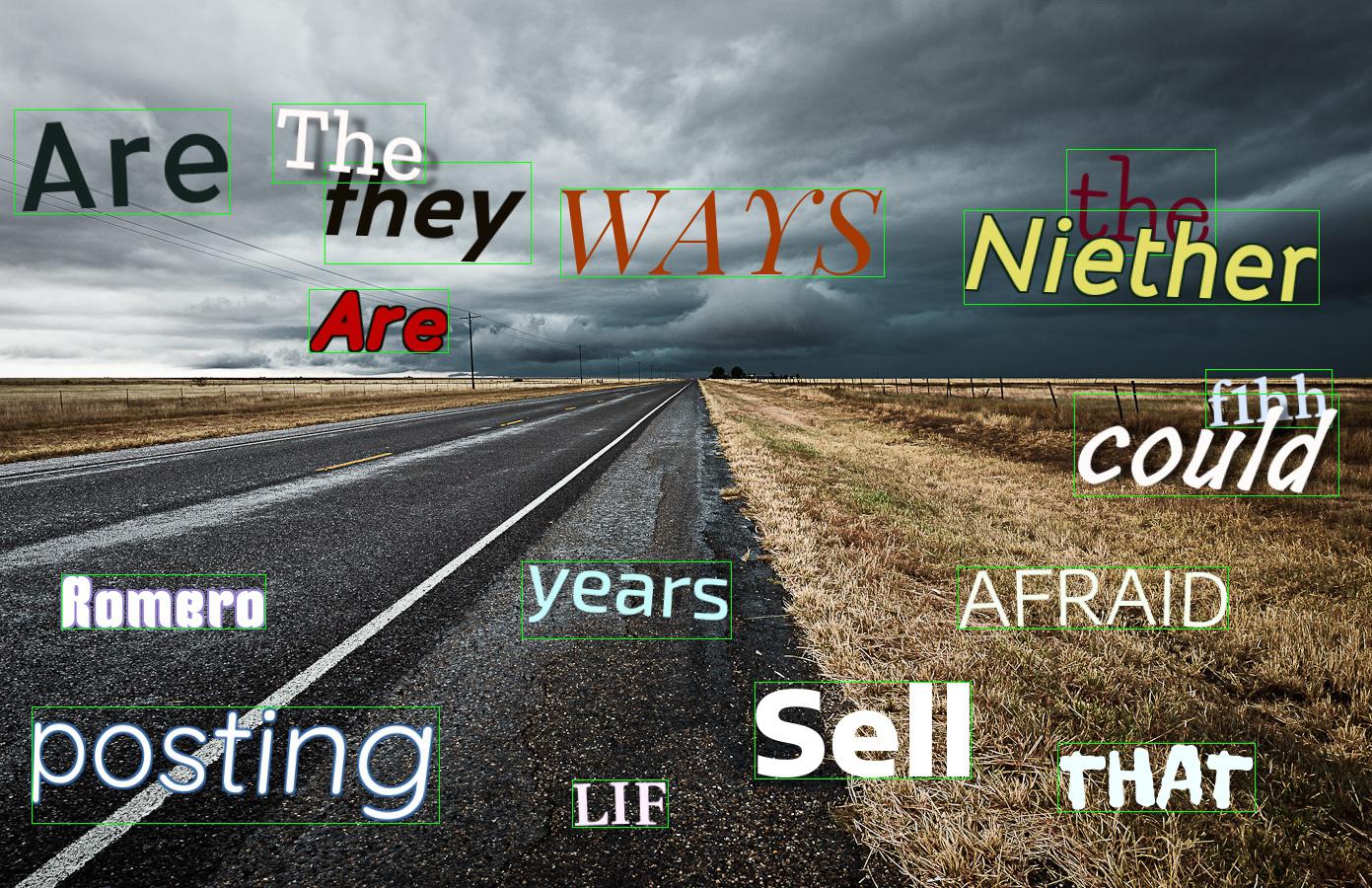}}
	\subfigure[]{
		\label{Fig.sub.16}
		\includegraphics[width=0.22\columnwidth,height=0.08\textheight]{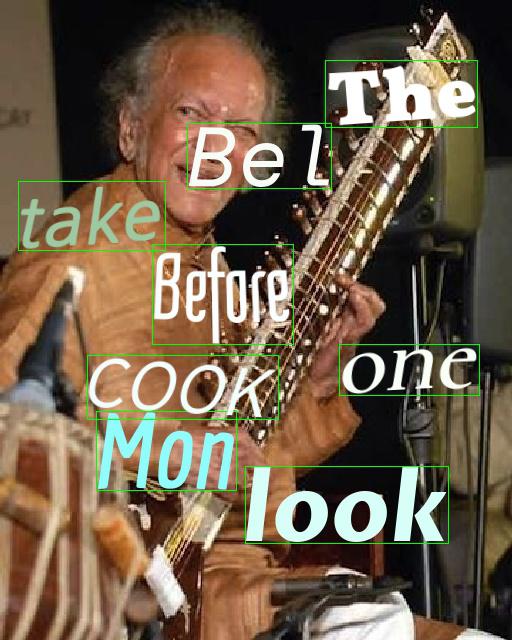}}
	\caption{Two problems ignored by existing methods. The generated texts colored by an unreasonable color scheme are shown in (a) to (d), which is hard to distinguish the texts and the background. (e) to (h) refer to the situation where text appears across depths in the real world. (i) to (l) are the results of some RGB images generated by our method, which successfully overcome the above two problems}
	\label{fig1}
\end{figure}
example,~\cite{b6} adopted the color scheme learned from the IIIT5K~\cite{b39} dataset. As shown in (a) to (d) of Fig. \ref{fig1}, the four images
are sampled from the dataset generated by~\cite{b6} itself.
The texts are framed in the figure, and it is not difficult to find that the color of the generated text is similar to the background color, which leads to the situation that the human eyes can not recognize. This example also proves the irrationality of the existing color scheme. Second, existing methods ignored the situation where text appears across depths in the image. One example is shown in Fig. \ref{fig1} (e) to (h), where (e) and (g) are real scene pictures,  (f) and (h) are the depth maps corresponding to these two pictures.

In this paper, we have designed a novel color scheme. The proposed method is based on the characteristics of the human eyes to observe things. Specifically, the human eye is insensitive to slight changes in the gray value. When the difference between the gray value of the text and the background is weak, it will cause the problem that the human eyes can not recognize it. Our method takes advantage of this feature for text coloring. Even when the text spans depth, the algorithm can analyze the color suitable for the current background. Some related examples have been shown in Fig.1 (i) to (l). For a fair verification, we take the same image mentioned in the above example as our background. The position of the text is randomly selected. In the algorithm, since we did not do any collision detection operation, there are some overlaps between texts. However, from the details in the pictures like (i), (k), we can find that the colors among the texts do not interfere with each other. It also proves the effectiveness of our method from the side.

The contributions of this paper can be summarized as follows: (1) We propose a new color scheme generation method based on human eye characteristics, which can avoid the negative effects caused by dirty data and improves the performance of the recognizers. (2) Text can appear in most positions of any picture, even across depths (3) The proposed method can replace the existing color scheme and help to do further research.

\section{Related Work}
\subsection{Human Eyes Characteristic} \label{Hec}

The visual information exchange between humans and machines is realized by generating image information on the screen and the transmission of image information is finally received by the human visual system. The visual system of our person is insensitive to small changes in grayscale, that is to say, in a grayscale image, when the grayscale value changes by one or two pixels, the human eyes can not distinguish it. According to~\cite{b12}, the human eyes in the image grayscale display system represented by 0$\sim$255, when the gray level is 8, 16, 32, the correct recognition rate of the human eyes is about 93.16\%, 68.75\%, 45.31\%. The gray level in the gray image represents the difference between bright pixels and dark pixels. When the gray level is higher, the degree of subdivision of gray pixels is greater, and on the contrary, the accuracy of human eye recognition is lower. The experimental results are in agreement with the actual situation. Moreover, this literature also mentioned the view that there is no significant difference in the resolution of the human eyes for grayscale images and RGB images. Our solution takes full advantage of this feature to filter out some pixels that the human eye can not distinguish from the current background for the best results.

\subsection{Synthesis Images}
The training of most deep learning algorithm models highly depends on large amounts of high-quality data. The advent of synthetic data successfully solved the problem of data shortage in real datasets. More importantly, the strategy of synthetic data as training data for the model and the real dataset for evaluation has been widely adopted by most current researches\cite{b5}.

Various image synthesis techniques have been applied to different deep learning tasks, such as object detection~\cite{b30,b33,b36}, scene understanding~\cite{b31,b34,b37}, human face~\cite{b3,b19,b35}, etc. For scene text detection and recognition tasks, synthetic data also made outstanding contributions. To train the scene text recognizer,~\cite{b11} proposed a text generation engine with the following steps to generate images: font rendering, shadow rendering, coloring, projection distortion, natural blending, and finally adding noise to simulate the environment.~\cite{b2} proposed to adopt GAN~\cite{b16} for generating scene text images. 

In the scene text detection task,~\cite{b6} adopted an innovative method on text location to produce data. It aimed to find the most suitable position for the text under the condition of the same depth and then used the color matching scheme learned from the IIIT5K dataset to colorize. To ensure semantic coherency was the highlight of~\cite{b15} and~\cite{b2}. They filtered the unreasonable positions such as the human face to select the position of the text, then finally adjusted the text through two steps of saliency analysis and adaptive coloring.~\cite{b7} and~\cite{b13} built a 3D world based on Unreal Engine 4 (UE4) to simulate the real-world. The greatest advantage of this method was that the whole scene was considered when rendering text and made the difference between the text instance and the background smaller.

Although these methods have shown excellent performance in synthesis technology, the existing color scheme they use for text coloring is unreasonable and has a certain negative effect on improving the performance of the model. At the same time, there are examples of text appearing across depth in reality which existing methods have ignored. Although~\cite{b7} and~\cite{b13} based on the 3D method can theoretically achieve the effect of text across depth by changing the viewing angle, the existing color scheme is difficult to support to get a clear and visible text.
\section{Methodology}
In this section, we will introduce our color scheme in detail. First of all, we describe the pipeline of the entire algorithm. Followed by that is the description of the two key thresholds in the algorithm. At last, we will present the part of text rendering.
\begin{figure*}[htbp]
	\centerline{\includegraphics[width=1\textwidth,height=0.25\textheight]{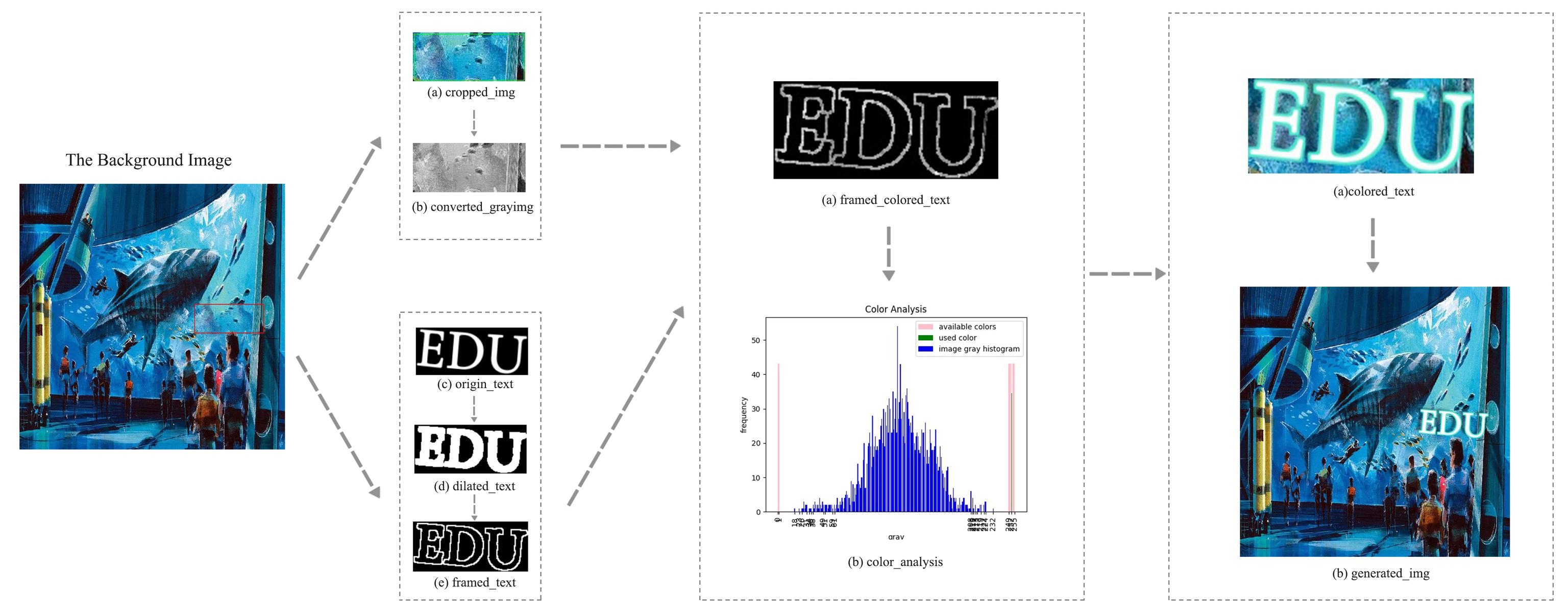}}
	\caption{(Left to right): (1) RGB image with a randomly selected text position. (2) Two parts: Cropped RGB image convert to grayscale image (or the original grayscale image); Get the text border by subtracting the original image from the dilated image. (3) Combine grayscale image and text, then analyze the distribution of the background image pixels and calculate the most suitable text color. (4) Revert text to RGB image and render it into the background.}
	\label{fig2}
\end{figure*}

\subsection{Overall Pipeline}
The process of generating data by our method is depicted in Fig. \ref{fig2}. In the preparation stage, we have sampled some text
\begin{figure}[htbp]
	\centerline{\includegraphics[width=0.9\columnwidth,height=0.2\textheight]{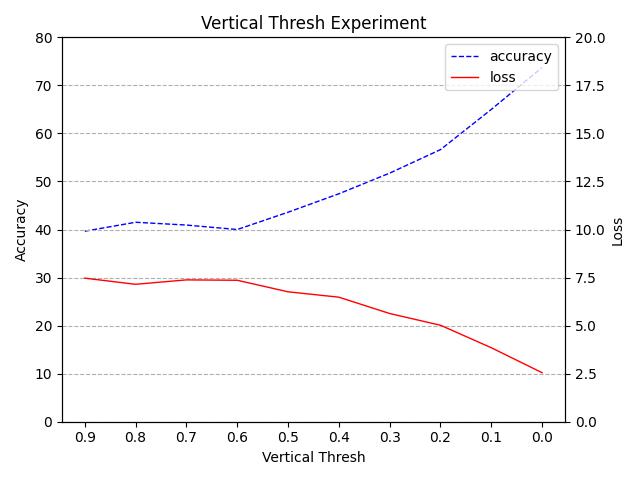}}
	\caption{Accuracy and Loss on the evaluation set under the different vertical threshold.}
	\label{fig3}
\end{figure}
fonts and background images. The font is extracted from Google fonts, which contains 3099 available Latin fonts. To verify the effectiveness of the proposed method, we randomly selected 100 images from the dataset collected by~\cite{b6} as our background images. The first column in Fig. \ref{fig2} is a randomly selected image, and then we randomly choose a position in the image for synthesizing text, whose size is the same as the text. There are two parts in the second column of Fig. \ref{fig2}. The first part is to crop the selected position and convert it into the corresponding grayscale image by taking the average value (if it is originally a grayscale image, this step could be omitted). The purpose of the second part is to get the border of the text. First, we randomly select a font and then add some operations such as rotation transformation. Secondly, we will expand the text by 2 pixel. Finally, we subtract the original text from the expanded text to get the text border. The third column presents the core part of our algorithm. We combine the border of the text obtained in the previous step with the converted grayscale background image to get the background color around the text. Then, we will analyze all the gray values of the background contained in the text border by our algorithm to get the most suitable text color. The details of this part will be described in the ensuing section. The final column shows the text colored by our algorithm and the generated image. It can be seen from the color analysis chart that the background pixels are complex, but our algorithm avoids the dirty data, and the resulting text is clear and visible.

\subsection{Thresholds}
This module is the core of our method and contains two parts. The first one (vertical threshold) is to determine some valid pixels that could be used for text coloring. The second (horizontal threshold) aims to select some available colors as candidate text colors from the first part.

\subsubsection{Valid \& Invalid Pixels.}
As shown in Fig. \ref{fig1} (a) to (d), the pixel used for text coloring makes a negative impact on the result of generating images. We call these types of pixels as invalid pixels. This section aims to acquire valid pixels from any input images, and this is also the first step of our algorithm in choosing the text color. Here is an example
that has been shown in the third column of Fig. \ref{fig2} (see color analysis chart), which illustrates the result of the text color chosen by two thresholds. In this color analysis chart, the ordinate represents the frequency of the color, and the abscissa represents the gray value from 0 to 255. First, we calculate the background color histogram according to the text border-image obtained in the previous step. Then, we filter out the background pixels through the threshold we set to get valid pixel values. Finally, we transfer the obtained results to the second threshold for analysis.

To get the best threshold for the algorithm, we make a comparison experiment. In the experiment, we uniformly adopt 50,000 data and 20 epochs of training and then test them on the IC03 dataset. As is shown in Fig. \ref{fig3}, we arrange the abscissa according to the percentage of the pixel frequency, the ordinate is the experimental evaluation which includes two parts: accuracy and loss. We can find that the two curves of ``accuracy" and ``loss" have a smooth trend, and finally achieve the best results at ``0". Therefore, our
first threshold is set to ``0", and the meaning of ``0" is to list all unused pixels. It is worth mentioning that our algorithm runs under the premise of two thresholds. Thus, in this experiment, our second threshold is set to ``0", and the function of the second threshold is to select available pixels,  when the value here is set to ``0", all pixels selected by the first threshold are allowed as candidate text colors.

More details about the first threshold are summarized in Algorithm \ref{a1}. In the pseudocode, ``hist" is used to receive the histogram result of the input gray image. ``maxValue" is the intermediate variable used to calculate the maximum pixel frequency in the histogram. ``thresh" represents the threshold in the algorithm. The output is to find out all the pixel indexes less or equal than this threshold.
\begin{algorithm}[htbp]
	\caption{Algorithm Of The Vertical Threshold}
	\label{a1}
	\begin{algorithmic}[1]
		\Require grayImg
		\Ensure unusedGrays
		\Function{getUnusedGrays}{grayImg,verticalThresh} 
		\State $verticalThresh \gets 0$
		\State $hist \gets the\ gray\ histogram\ of\ grayImg$
		\State $maxValue \gets max(hist)$  
		\State $thresh \gets maxValue * verticalThresh $
		\State $unusedGrays \gets all\ index\ array\ of\ hist\ which$  
		\Statex$\qquad \qquad frequency\ is\ less\ or\ equal\ than\ thresh$		
		\State \Return {unusedGrays} 
		\EndFunction
	\end{algorithmic}
\end{algorithm}

\subsubsection{Suitable Candidate Color.}
\begin{figure}[htbp]
	\centerline{\includegraphics[width=0.9\columnwidth,height=0.2\textheight]{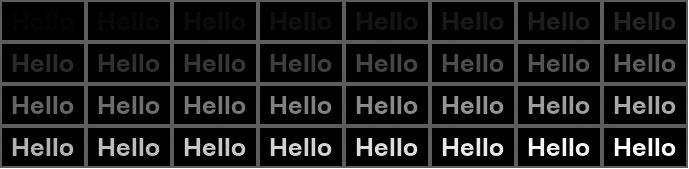}}
	\caption{The performance of gray pixel differences under the condition of 32-level gray values. }
	\label{fig4}
\end{figure}

This section is based on the previous content, and the purpose is to choose the available part from the obtained pixels as the candidate for the final text coloring. The setting of the second threshold references the characteristics of human eyes to observe things. To show these characteristics more clearly, we have prepared an example, as is shown in Fig. \ref{fig4}. This image explicitly describes the process of the gradation of gray values. Here we use black as the background for a better visual effect. We choose ``hello" as the text content, and it will serve as a carrier to show us the differences between different pixels. There is a total of 256 gray pixels, and we divide them into 32-level gray value, which means that each rectangle in the figure represents 8-pixel value. As mentioned above, the recognition accuracy of the human eyes corresponding to the 32-level gray value is about 45.31\%.  From the picture, we can find that the text under the adjacent pixels is difficult to distinguish, which is also the limitation of our human eyes.

We combine the literature~\cite{b12} to draw such a conclusion--when the interval pixel is larger (means that the pixel-level is smaller), the recognition accuracy of the human eyes will be better. As is shown in Fig. \ref{fig5}, to prove our conjecture, we make an experiment on different interval pixels. For ensuring the rigor of the experiment, we also use 0 and 8 as the interval pixel. Note that we use ICDAR2003~\cite{b8} as the validation dataset and most of the text appears in the same depth. The background is simple and has more pixels available. 

After analyzing the accuracy and loss map, we finally decide to utilize 16 as our second threshold. Two reasons for that: (1) Although smaller thresholds such as 0 and 8 also perform well, the text background in the validation dataset is relatively simple. Once the text crosses the background depth, the color scheme selected by this type of threshold is likely to become dirty data, which will affect the results of the experiment. It can also be confirmed from the loss graph, where the performance of the smaller threshold is not as well as the larger. (2) In principle, the larger the threshold, the better the performance. However, there exists such a problem: when the background pixels are too complicated and the interval pixels are also too large, which may result in no available
pixels. Moreover, in the accuracy graph, the larger threshold like 24 doesn't achieve an excessive advantage.

As just mentioned, to avoid the problem of no usable pixels, we have adopted this strategy: if there are no usable pixels after algorithmic analysis, it will return to the first step to reselect a text position and analyze again. We set the number of iterations for 20 rounds. Once this number is exceeded, this synthesis will be abandoned.
\begin{algorithm*}[ht]
	\caption{Algorithm Of The Horizontal Threshold}
	\label{a2}
	\begin{algorithmic}[1]
		\Require grayImg, minMargin
		\Ensure colors
		\Function{getDesignColors}{grayImg, minMargin=16}  
		\State $unusedGrays \gets call\ getUnusedGrays$(grayImg)$ $
		\algorithmiccomment{Receive the result of the previous algorithm}
		\If{minMargin $\leq$ 0}
		\State \Return {unusedGrays}
		\EndIf
		\State \bm{$//$} \bm{$find$} \bm{$all$} \bm{$edgecolors$} 
		\State $edgeColors \gets [] $
		\State $length \gets the\ length\ of\ unusedGrays$
		\State $concatGrays \gets [-1, unusedGrays[0...length], 256] $
		\State $lastRightG \gets -1$
		\For{$i \gets 1\ to\ length\ +\ 1$}
		\State $leftG \gets concatGrays[i-1]$
		\State $currentG \gets concatGrays[i]$	
		\If{$currentG\ -\ leftG\ !=\ 1\ and\ currentG\ -\ 1\ !=\ lastRightG$}
		\State $edgeColors.append(currentG\ -\ 1)$
		\EndIf
		\State $rightG\ =\ concatGrays[i\ +\ 1]$
		\If{$rightG\ -\ currentG\ !=\ 1$}
		\State $lastRightG\ =\ currentG\ +\ 1$
		\State $edgeColors.append(lastRightG)$
		\EndIf
		\EndFor
		\State \bm{$//$} \bm{$give$} \bm{$up$} \bm{$colors$} \bm{$surrounding$} \bm{$edgeColors$} \bm{$by$} \bm{$margin$} \bm{$color$}
		\State $colorValidInfo \gets [0,0,0...0]\ which\ length\ is\ 256 $
		\For{$g\ in\ unuseGrays$}
		\State $colorValidInfo[g]\ =\ 1 $
		\EndFor
		\For{$c\ in\ edgeColors$}
		\State $lIndex\ =\ max(0,\ c\ -\ minMargin)$
		\State $rIndex\ =\ min(256,\ c\ +\ minMargin\ +\ 1)$
		\State $colorValidInfo[lIndex...rIndex]\ =\ 0$ \algorithmiccomment{Left closed interval and right open interval}
		\EndFor
		\State \bm{$//$} \bm{$summary$} \bm{$all$} \bm{$available$} \bm{$colors$}
		\State $colors \gets 
		all\ index\ array\ of\ colorValidInfo\ which\ value\ is\ 1 $
		\State \Return {colors} 
		\EndFunction
	\end{algorithmic}
\end{algorithm*}

The details of the second threshold are summarized in Algorithm \ref{a2}. There are three parts in the pseudocode: ``Find the edge-color", ``Filter the interfering pixels", ``Summarize all available pixels". ``unusedGrays" receives all the unused pixels obtained in the previous algorithm. In the first part, ``edgeColors" is a list, ``concatGrays" is a list after
connecting ``unusedGrays" with -1 and 256. In the second part, ``colorValidInfo[lIndex...rIndex] = 0'' means all pixels within the threshold range more or less than the ``edgeColors" will be filtered out. The last part is to summarize all the surviving pixels.


\subsection{Text rendering}
From what has been mentioned above, we know that in the first threshold we finally adopt the strategy of extracting all unused pixels, namely, the 0 and we choose 16 with the best performance as the second threshold. Subsequently, we will introduce the part of text rendering in this section.

\subsubsection{Gray Image \& RGB Image.}
The synthesis process of grayscale images is roughly same as RGB images. The only difference is the second and fourth steps, where RGB and grayscale need to be converted to each other. The principle to choose the final pixel from the candidate is randomly selected. For a fair comparison, we choose Google Fonts as these methods~\cite{b16,b13} do, and the same text corpus as~\cite{b11}. Thanks to this novel color scheme, text coloring can avoid the negative effect caused by dirty data.

\section{Evaluation}
In this section, we make some evaluations in the following parts: ``text recognition'', ``color scheme'', and ``synthesis speed''. The first part will introduce some representative datasets used in our experiments. The second part describes some details and settings of each experiment. The third part makes a comparison experiment on text recognition between some synthesis algorithms and recognition algorithms. The fourth part compares our color scheme with the existing one. The last one presents the speed of our algorithm to synthesize images.	

\subsection{Datasets}\label{datasets}
The benchmark datasets for text recognition evaluation: ICDAR 2003 (IC03), ICDAR 2013 (IC13), Street View Text (SVT), Synth90K, SynthText, VISD. These datasets are elaborated in the next.

\subsubsection{IC03~\cite{b8}.}
This is a regular dataset, which contains 251 images for testing and 867 cropped text instances.

\subsubsection{IC13~\cite{b10}.}
This dataset is similar to the IC03 and much of the data is inherited from it. It contains 1015 ground-truth cropped images in total.

\subsubsection{SVT~\cite{b40}.}
The dataset is sampling from the street images which is both more complex and challengeable than ICDAR datasets. It contains 250 test images and 647 cropped images.

\subsubsection{Synth90K~\cite{b11}.}
This dataset contains 9,000,000 text instance images. Each image is annotated with a ground-truth label.

\subsubsection{SynthText~\cite{b6}.}
The dataset contains 8,000,000 text training images. The innovation of producing data is on choosing the position. 

\subsubsection{VISD~\cite{b15}.}
The innovation of this dataset is to make semantic coherence. It filters the disharmonious region like the human face to place text. It contains 5,000,000 text instance images at all.

\subsection{Implementation Details}\label{ID}

\subsubsection{Recognition Model.}
The recognition model we used CRNN~\cite{b4}. Different from other models, the input image adopts the whole image and the CTC~\cite{b14} as loss function, which makes the model capable to deal with some sequence problems. More importantly, it can handle sequence recognition problems of any length, no need for pre-defined lexicons.

\subsubsection{Details.}
The first experiment is text recognition. The objects for comparison are some state-of-the-art synthesis methods and some recognition algorithms. To create a fair comparison environment, we all use the same amount of training data. In the second experiment, we have compared the existing color scheme like IIIT5K. We note that the coloring scheme of~\cite{b6} is learned from it, and some subsequent literature also refers to this form of coloring scheme. The last experiment is the evaluation of the speed of generating images. All experiments are implemented on an ubuntu workstation with an 8-core Intel CPU, an NVIDIA TITAN RTX GPU, and 16G RAM. Our method omits the step of finding the most suitable position for text(including the conditions of the same depth, etc.)
\begin{figure*}[ht]
	\centering
	\subfigure[]{
		\label{Fig.sub.17}
		\includegraphics[width=0.4902\textwidth,height=0.25\textheight]{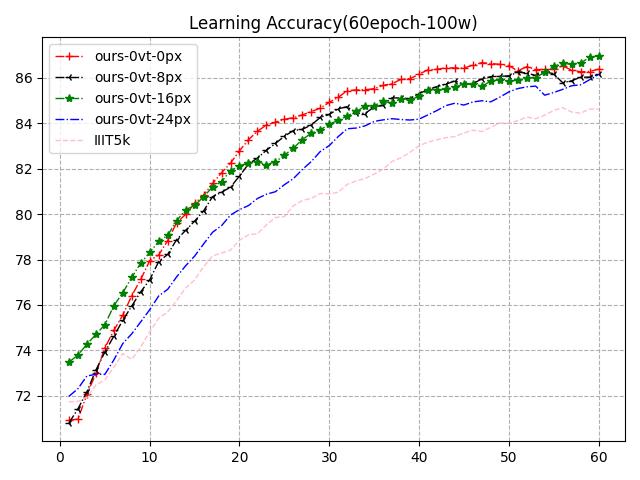}}
	\subfigure[]{
		\label{Fig.sub.18}
		\includegraphics[width=0.4902\textwidth,height=0.25\textheight]{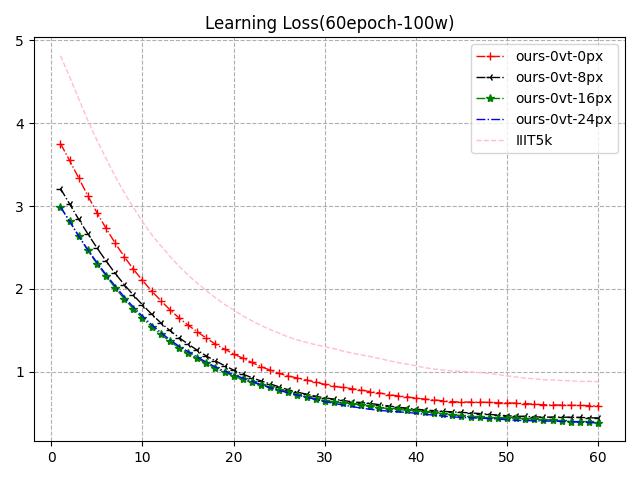}}
	\caption{The comparison of color schemes between the IIIT5K dataset and ours.}
	\label{fig5}
\end{figure*}
\begin{figure}[ht]
	\centerline{\includegraphics[width=0.9\columnwidth,height=0.2\textheight]{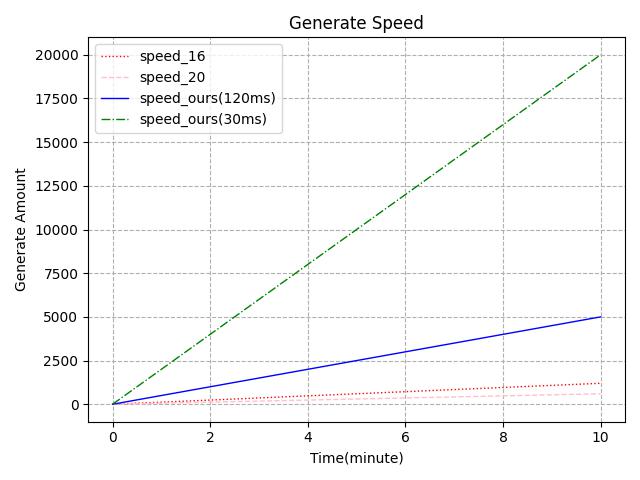}}
	\caption{The speed comparison between different methods, ``speed\_16" refers to ~\cite{b6}, and ``speed\_20" refers to ~\cite{b13}. ``speed\_ours(30ms)" represents that generating an image with ten words costs 30ms. The same explanation is for ``speed\_ours(120ms)".}
	\label{fig6}
\end{figure}

\subsection{Experiments On Text Recognition}\label{EOTR}
The proposed synthesis technique is evaluated over three public
datasets including ICDAR 2003, ICDAR 2013, SVT as shown in table \ref{Tab1}. In the table, our CRNN model has achieved state-of-the-art text scene recognition accuracy. In the absence of a lexicon, its performance is better than other methods, especially on the IC03 dataset, the accuracy has reached 92.8\%, where is 2.7 points more than the best performance. Similarly, on the IC13 dataset, it is 2.5 points higher than the best performance. It should be noted that its performance on SVT is worse than other methods. The reason is that we only do some simple operations to generate curve text, and its requirements are far from satisfying the model to predict more complex scene text. 

Table \ref{Tab1} also shows the comparison between our proposed method and the three existing synthesis methods. In particular, for a fairer comparison, we all adopt the CRNN model and use 5 million(5M) data to train the model. It is not hard to find that our method performs better than other methods on both two datasets. The poor performance of other methods is largely due to the unreasonable text coloring scheme, which is also the point that these methods have ignored.

\subsection{Color Scheme Evaluation}\label{CSE}
The unreasonable statement about the existing color scheme has been mentioned in some of the above subsections. More details will be presented in this paragraph. We take IIIT5K as an example and compare it with our methods under different thresholds. The experimental setting includes 1 million training data, 60 training epochs, and IC03 as the verification dataset. Moreover, we use 0.9 as the curve smoothing coefficient. Here we use less training data(1 million), which causes the situation that the performance in this figure is worse than it shows in the above table. 
\begin{table}[ht]
	\scriptsize
	\centering
	\caption{Comparison between previous methods and ours on the scene text recognition.}
	\begin{tabular}{lccc}
		\toprule
		\multirow{2}{*}{Methods} & 
		\multicolumn{1}{c}{ICDAR2003} & \multicolumn{1}{c}{ICDAR2013}  & 
		\multicolumn{1}{c}{SVT} \\
		\cmidrule(r){2-4} 
		&  None
		&  None 
		&  None       \\
		\midrule
		Mishra\ et\ al. ~\cite{b25}                           
		& 67.7         
		& 60.2        
		& -  \\
		Yin\ et\ al. ~\cite{b26}                            
		& 81.1         
		& 81.4        
		& 72.5  \\
		PhotoOCR ~\cite{b27}                             
		& -         
		& 87.6        
		& 78.0  \\
		STAR-Net~\cite{b24}                         
		& 89.9         
		& 89.1        
		& \bm{$83.6$}  \\
		R${^2}$AM ~\cite{b23}                             
		& 88.7         
		& 90.0        
		& 80.7  \\
		CRF~\cite{b29}                             
		& 89.6         
		& 81.8        
		& 71.7  \\
		RARE~\cite{b28}                             
		& 90.1         
		& 88.6        
		& 81.9  \\
		\midrule
		CRNN\ +\ 90K(5M) ~\cite{b11}             
		& 89.4           
		& 86.7        
		& 80.8         \\
		CRNN\ +\ ST(5M)~\cite{b6}             
		& -           
		& 86.4   
		& 79.2\\
		CRNN\ +\ VISD(5M) ~\cite{b15}             
		& -         
		& 87.1     
		& 81.5         \\
		\midrule
		CRNN\ +\ Ours(5M)             
		& \bm{$92.8$}         
		& \bm{$92.5$}   
		& 80.7            \\
		\bottomrule
	\end{tabular}
	
	\label{Tab1}
\end{table}

As shown in Fig. \ref{fig5}, the pink dotted line represents the performance of IIIT5K, and other lines represent our method. Whether it is an accuracy or loss graph, the performance of IIIT5K is worse than ours. The reason for this is because of the dirty data in its dataset, and its existence cause the problem of confusion between text and background . Since our method aims to avoid such dirty data, our method performs better than IIIT5K even under different threshold conditions.

Moreover, we find that the irrationality of the existing color scheme is also manifested in the fact that when the text crosses the depth, the last available color pixel overlaps the background color pixel(we provide this part in the supplementary materials). This is because this type of color scheme takes the strategy of taking the average, and once the background is complicated, the final color pixel obtained by the algorithm is likely to overlap with the background pixels. After rendering the text with this color, it may cause the confusion problem, and this fixed key-value pair method provides fewer candidate colors. Meanwhile, the distribution of available colors is also denser. Different from it, the result obtained through our scheme has a wider color distribution and does not overlap with the background color. It provides more possibilities for text rendering. Some synthetic samples have proven the effectiveness of this scheme.

\subsection{Speed Evaluation}\label{SE}
In this section, we will verify the superiority of our algorithm's speed. As can be seen from the Fig. \ref{fig6}, the abscissa represents time, in minutes, and the ordinate represents the number of generating pictures. We use lines of different colors to indicate the speed of each algorithm to synthesize pictures. The two dashed lines with pink and red color represent the speed of the algorithm~\cite{b13} and the algorithm~\cite{b6}, and the other two lines represent our method. In the figure,~\cite{b13} generates about 40 to 60 pictures every minute, and~\cite{b6} generates about 120 pictures in one minute (The facts are mentioned by themselves). After a rough calculation, the speed of our algorithm to generate a picture containing only one text is about 3 milliseconds. Here we assume a picture containing 10 words, which means that the speed is 30 milliseconds to produce a picture. Similarly, we recorded the speed of 120 milliseconds. From the image, easy to see that the steepness of the slope of our algorithm is larger and this also verfies the advantage of the speed of our algorithm mentioned above.

\section{Conclusions}
This paper introduces a new way of thinking about data synthesis. We mainly start of the low-level feature of the text--text color,  taking into account the limited grayscale that the human eye can distinguish, and with this visual feature, we designed a color scheme generation method. The colored text can be easily distinguished from the complex background. More importantly, our method allows text to appear in the image across depths. Therefore, our method is helpful to improve the performance of the model. However, there is still much work worthy of in-depth study. For example, in this paper, we only make the recognition experiment, and in principle, the performance of the detector can also be improved through our method. In the next step, we will conduct experiments specifically for detection tasks. Although we don't need to analyze the best placement of the text, our color analysis is a bit time-consuming, which is a disadvantage compared to the color scheme learned directly from the dataset. Thus, how to reduce the complexity of the algorithm is also a problem that we need to consider in the future.

\bibliographystyle{IEEEtran}
\bibliography{references}

\end{document}